\pdfoutput=1

\documentclass[11pt]{article}

\usepackage[]{acl}

\usepackage{times}
\usepackage{latexsym}

\usepackage[T1]{fontenc}

\usepackage[utf8]{inputenc}

\usepackage{microtype}

\usepackage{pbox}
\usepackage{makecell}
\usepackage{multirow}
\usepackage{graphicx}

\usepackage{soul}
\sethlcolor{lightgray}

\definecolor{ultramarine}{rgb}{0.07, 0.04, 0.56}

\definecolor{mypink1}{rgb}{0.858, 0.188, 0.478}
\definecolor{mygreen}{rgb}{0.258, 0.888, 0.178}

\usepackage{xcolor,pifont}
\newcommand*\colourcheck[1]{%
  \expandafter\newcommand\csname #1check\endcsname{\textcolor{#1}{\ding{51}}}%
}
\newcommand*\colouruncheck[1]{%
  \expandafter\newcommand\csname #1uncheck\endcsname{\textcolor{#1}{\ding{53}}}%
}
\colourcheck{green}
\colouruncheck{red}
\colouruncheck{blue}

\usepackage{booktabs}

\title{How Well Do Multi-hop Reading Comprehension Models Understand Date Information?}

\author{
Xanh Ho,$^{\diamondsuit, \heartsuit}$
Saku Sugawara,$^\heartsuit$\and
Akiko Aizawa$^{\diamondsuit, \heartsuit}$ \\
$^\diamondsuit$The Graduate University for Advanced Studies, Kanagawa, Japan\\
$^\heartsuit$National Institute of Informatics, Tokyo, Japan \\
\{xanh, saku, aizawa\}@nii.ac.jp
}

\begin{document}
\maketitle

\begin{abstract}

Several multi-hop reading comprehension datasets have been proposed to resolve the issue of reasoning shortcuts by which questions can be answered without performing multi-hop reasoning.
However, the ability of multi-hop models to perform step-by-step reasoning when finding an answer to a comparison question remains unclear. It is also unclear how questions about the internal reasoning process are useful for training and evaluating question-answering (QA) systems. To evaluate the model precisely in a hierarchical manner, we first propose a dataset, \textit{HieraDate}, with three probing tasks in addition to the main question: extraction, reasoning, and robustness. Our dataset is created by enhancing two previous multi-hop datasets, HotpotQA and 2WikiMultiHopQA, focusing on multi-hop questions on date information that involve both comparison and numerical reasoning. We then evaluate the ability of existing models to understand date information.
Our experimental results reveal that the multi-hop models do not have the ability to subtract two dates even when they perform well in date comparison and number subtraction tasks. Other results reveal that our probing questions can help to improve the performance of the models (e.g., by +10.3 F1) on the main QA task and our dataset can be used for data augmentation to improve the robustness of the models.


\end{abstract}

\section{Introduction}

\begin{figure}[htp]
    \includegraphics[scale=0.6921]{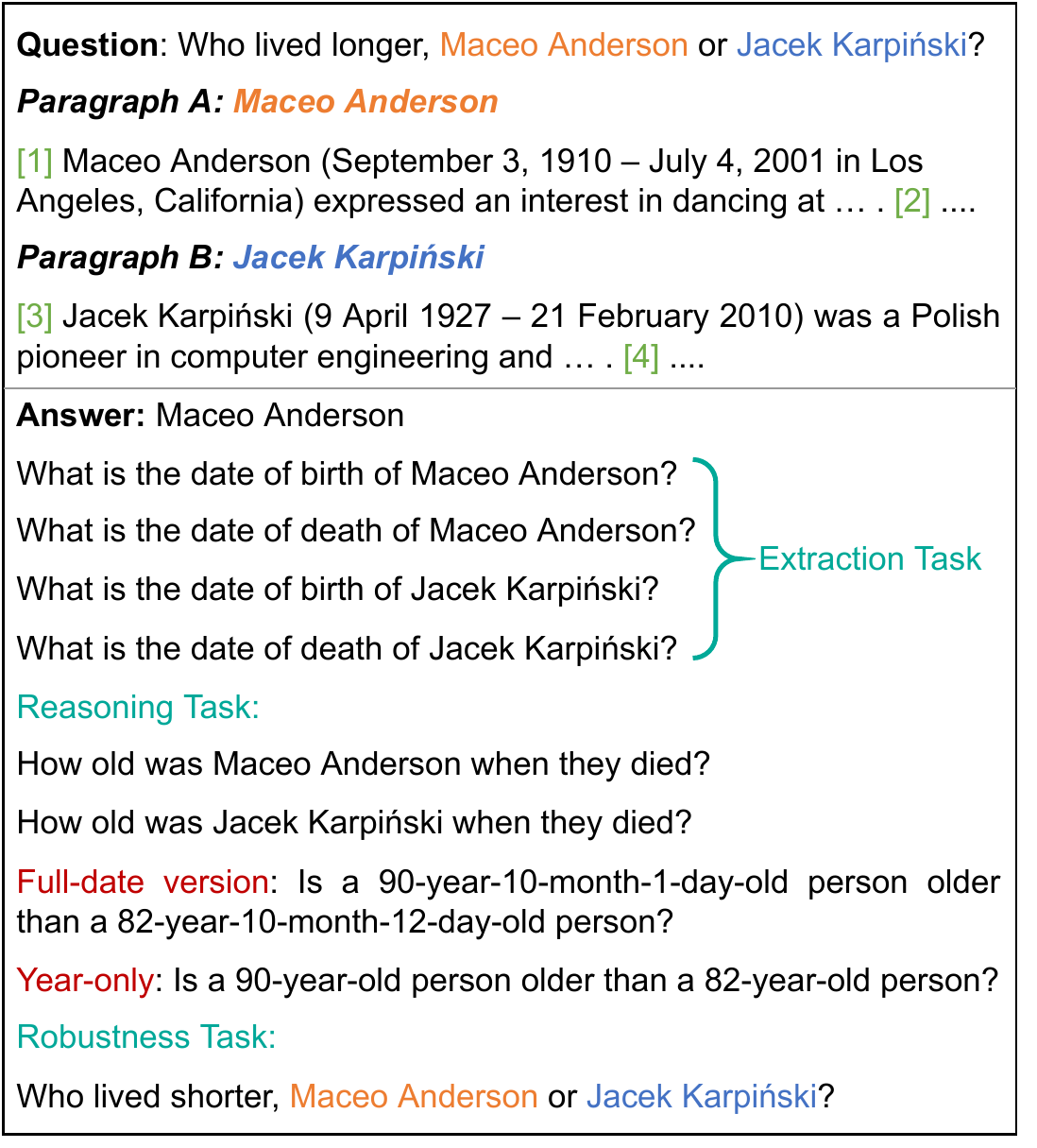}
    \caption{Example of a question in our dataset.}
    \label{fig:example_combine}
\end{figure}
\renewcommand{\floatpagefraction}{.9}

Multi-hop reading comprehension (RC) requires a model to read and aggregate information from multiple paragraphs to answer a given question~\cite{welbl-etal-2018-constructing}.
Several datasets have been proposed for this task, such as HotpotQA~\cite{yang-etal-2018-hotpotqa} and 
2WikiMultiHopQA~\citep[2Wiki;][]{ho-etal-2020-constructing}.
Although the proposed models show promising performances, previous studies ~\cite{jiang-bansal-2019-avoiding,chen-durrett-2019-understanding,min-etal-2019-compositional,tang-etal-2021-multi} have demonstrated that existing multi-hop datasets contain reasoning shortcuts, in which the model can answer the question without performing multi-hop reasoning.

There are two main types of questions in the previous multi-hop datasets: bridge and comparison.
\citet{tang-etal-2021-multi} explored sub-questions in the question answering (QA) process for model evaluation. 
However, they only used the bridge questions in HotpotQA and did not fine-tune the previous multi-hop models on their dataset when performing the evaluation.
Therefore, it is still unclear about the ability of multi-hop models to perform step-by-step reasoning when finding an answer to a comparison question. 


%
HotpotQA provides sentence-level supporting facts (SFs) to explain the answer.
However, as discussed in \citet{inoue-etal-2020-r4c} and \citet{ho-etal-2020-constructing}, the sentence-level SFs cannot fully evaluate
the reasoning ability of the models;
to solve this issue, in addition to sentence-level SFs, these studies provide a set of triples as the evidence information.
For example, for the question in Figure~\ref{fig:example_combine}, the evidence regards the dates of birth and death of two people, 
e.g., \textit{(Maceo, date of death, July 4, 2001)}.
We argue that simply requiring the models to detect a set of triples, in this case, cannot explain the answer to the question and cannot describe the full path from the question to the answer;
additional operations, including calculations and comparisons, need to be performed to obtain the final answer.
%

To deal with this issue, we introduce a dataset, \textit{HieraDate},\footnote{Our data and code are available at \url{https://github.com/Alab-NII/HieraDate}.} consisting of the three probing tasks. 
(1) The extraction task poses sub-questions that are created by converting evidence triples into natural language questions.
(2) The reasoning task is pertinent to the combination of triples, involving comparison and numerical reasoning that precisely evaluate the reasoning path of the main questions.
(3) The robustness task consists of examples generated by slightly changing the semantics (e.g., \textit{born first} to \textit{born later}) of the original main questions.
The purpose of the robustness task is to ensure that the models do not exploit superficial features in answering questions.

Our dataset is created by extending two existing multi-hop datasets, HotpotQA and 2Wiki.
As the first step of the proof of concept, we start with the date information through comparison questions because this information
is available and straightforward to handle.
Moreover, based on the classification of comparison questions in \citet{min-etal-2019-compositional}, all comparison questions on date information require multi-hop reasoning for answering.
We then use our dataset to evaluate two leading models, HGN~\cite{fang-etal-2020-hierarchical} and NumNet+~\cite{ran-etal-2019-numnet} on two settings: with and without fine-tuning on our dataset.
We also conduct experiments to investigate whether our probing questions are useful for improving QA performance and whether our dataset can be used for data augmentation.

Our experimental results reveal that existing multi-hop models perform well in the extraction and robustness tasks but fail in the reasoning task when the models are not fine-tuned on our dataset.
We observe that with fine-tuning, HGN can perform well in the comparison reasoning task; meanwhile, NumNet+ struggles with subtracting two dates, although it can subtract two numbers.
Our analysis shows that questions that require both numerical and comparison reasoning are more difficult than questions that require only comparison reasoning.
We also find that training with our probing questions boosts QA performance in our dataset,
showing improvement from 77.1 to 82.7 F1 in HGN and from 84.6 to 94.9 F1 in NumNet+.
Moreover, our dataset can be used as augmentation data for HotpotQA, 2Wiki, and DROP~\cite{dua-etal-2019-drop}, which contributes to improving the robustness of the models trained on these datasets.
Our results suggest that a more complete evaluation of the reasoning path may be necessary for better understanding of multi-hop models' behavior.
We encourage future research to integrate our probing questions when training and evaluating the models.

\section{Related Work}

In addition to~\citet{tang-etal-2021-multi}, \citet{al-negheimish-etal-2021-numerical} and \citet{geva2021break} are similar to our study.
\citet{al-negheimish-etal-2021-numerical} evaluated the previous models on the DROP dataset to test their numerical reasoning ability. 
However, they did not investigate the internal reasoning processes of those models.
\citet{geva2021break} proposed a framework for creating new examples using the perturbation of the reasoning path.
Our work is different in that their focus was on creating a framework, and it does not necessarily ensure the quality of all generated perturbation samples.
Moreover, we investigate the QA process in-depth, while \citet{geva2021break} do not include all detailed questions (e.g., they do not include extraction task and comparison reasoning questions in Figure 1).

\section{Dataset Construction}
\label{data_construction_main}

Our dataset is generated by using the two existing multi-hop datasets, HotpotQA and 2Wiki (more details are in Appendix~\ref{sec_appendix_data_previous}).

\paragraph{Obtain Date Questions}
We first sampled the comparison questions in HotpotQA and 2Wiki.
We then used a set of predefined keywords, such as \textit{born first} and \textit{lived longer}, 
to obtain questions regarding the date information.
%
From the train and dev. split, respectively, we obtained 119 (after annotating, only use 114 samples) and 878 samples in HotpotQA, and 984 and 8,745 samples in 2Wiki.

\begin{table}[t]
  \begin{center}
    \begin{tabular}{p{1.0cm}p{5.6cm}} 
     \toprule

      Task  & Templates/Phrases
    \\ 
      \midrule
    
  \multirow{10}{*}{Extract}  & What is the birth date of \#name?  \\ 
 & What's the birth date of \#name?  \\ 
 & What is the date of birth of \#name?  \\ 
 & What's the date of birth of \#name?  \\ 
 & When was \#name born?  \\ 

 & What is the death date of \#name?  \\ 
 & What's the death date of \#name?  \\ 
 & What is the date of death of \#name?  \\ 
 & What's the date of death of \#name?  \\ 
 & When did \#name die?  \\ 
       \midrule
       
   \multirow{7}{*}{Reason}  & Does \#date1 come before \#date2?  \\ 
 & Does \#date1 come after \#date2?  \\ 
 & How old was \#name when they died?  \\ 
   
 & Is a \#age1 person younger than a \#age2 person? \\
 
 & Is a \#age1 person older than a \#age2 person? \\
        \midrule

   \multirow{6}{*}{Robust}  & Born first/earlier $\Leftrightarrow$ Born later  \\
 
 & Born later  $\Leftrightarrow$ Born first  \\
 
 & Died first/earlier  $\Leftrightarrow$ Died later \\ 
  & Died later/second/last   $\Leftrightarrow$ Died first \\ 
  & Died more recently  $\Leftrightarrow$ Died first \\ 
 & Lived longer  $\Leftrightarrow$ Lived shorter \\ 
    \bottomrule

    \end{tabular}
    \caption{List of templates and phrases that we used in the dataset creation process. \textit{Extract}, \textit{Reason}, and \textit{Robust} represent the three tasks: extraction, reasoning, and robustness, respectively.
    }
    \label{tab:templates}
  \end{center}
\end{table}

\begin{table}[t]
  \begin{center}
    \begin{tabular}{l | r | r r r} 
    \toprule
 
    Split  & Main & Extract & Reason &  Robust \\

      \midrule
    Train  & 8745 & 21340 & 19415 & 8745  \\
    
    Dev.  & 549 & 1346 & 1222 & 549 \\
     
    Test  & 549 & 1346 &  1222 & 549 \\

    \bottomrule

    \end{tabular}
    \caption{Our dataset statistics. Each main question has the extraction, reasoning, and robustness tasks.
    }
    \label{tab_dataset}
  \end{center}
\end{table}

\paragraph{Generate Sub-questions and Sub-answers}
In 2Wiki, we used the evidence in the form of triples (e.g., (Maceo, date of death, July 4, 2001)) to automatically generate sub-questions and sub-answers for the extraction task.
We used Wikidata IDs (available in 2Wiki) to obtain structured date information to compare and/or subtract two dates when generating questions for the reasoning task. 
To obtain natural language questions, we wrote ten and five templates for the extraction and reasoning tasks, respectively.
%
Similar to~\citet{min-etal-2019-multi}, to evaluate the robustness of the models, we created the adversarial questions by changing the main multi-hop questions such that the new answers are opposite
(e.g., we changed the question: ``Who lived longer, A or B?'' to ``Who lived shorter, A or B?'').
We observed that the ten phrases (e.g., \textit{born first}) could cover all questions, and used these phrases to generate robustness questions.
Table~\ref{tab:templates} presents a set of templates and phrases that we used in the dataset creation process.

In HotpotQA, unlike 2Wiki, no triples are available; therefore, we first prepared triples for the sampled questions, and then performed the same procedure as in 2Wiki to 
generate all probing questions.
To obtain the triples, we first filtered the distractor paragraphs and retained only gold paragraphs.
We then used Spacy\footnote{\url{https://spacy.io/}} to extract the entities in the questions.
Further, we manually annotated the date with two formats: unstructured (e.g., `May 1992') and structured (e.g., month=5).
It is noted that we used only the dev. set in HotpotQA.

\paragraph{Construct HieraDate}
We created our dev. and test sets from the dev. sets of HotpotQA and 2Wiki, and our training set from the 2Wiki training set.
Table~\ref{tab_dataset} lists the number of samples for each task and each split in our dataset.
Our dataset includes two main types of questions: questions that ask about both date-of-birth and date-of-death information (e.g., ``who lived longer''), and those that ask about only the date-of-birth or date-of-death information (e.g., ``who was born later'').
We call the first type \textit{combined reasoning} because it requires both comparison and numerical reasoning (Figure~\ref{fig:example_combine}).
The second type is called \textit{comparison reasoning} 
(Figure~\ref{fig:example_born} is in Appendix~\ref{sec_appendix_data_analysis}) 
because it requires only comparison reasoning.
\textit{One combined reasoning} sample has one main multi-hop question, four extraction questions, two numerical reasoning questions, one comparison question, and one robustness question.
Meanwhile, \textit{one comparison reasoning} sample has one main multi-hop question, two extraction questions, two comparison questions, and one robustness question.

\begin{table*}[htp]
  \begin{center}
    \begin{tabular}{c l r r r r r r r r} 
     \toprule
     
    \multirow{2}{0.8cm}{Fine-tuning} & \multirow{2}{1.5cm}{Model} & \multicolumn{2}{c}{Main}  & \multicolumn{2}{c}{Extraction}  &
      \multicolumn{2}{c}{Reasoning} & 
      \multicolumn{2}{c}{Robustness} \\

      \cmidrule{3-10}
     &  & EM & F1 & EM & F1 & EM (num) & EM (comp) & EM & F1 \\
       \midrule

 {\multirow{2}{*}{\blueuncheck}}


  &  HGN  & 66.85 & 76.15 & 94.58 & 96.14 & \textit{N/A}  & 53.08 & 71.95 & 81.64 \\       
  
   

  &  NumNet+  & 67.94  & 71.57 & \hphantom{0}1.26 & 47.93 &   22.79 (F1)  & \textit{N/A} & 69.58 & 71.91 \\ 
  

    \midrule

     {\multirow{2}{*}{\greencheck}}
    &  HGN  & 78.87 & 82.69 & 96.06 & 97.14 &  \textit{N/A}  & 100 & 76.68 & 78.58 \\ 
    

  &  NumNet+  & 95.08 & 95.20 & 96.36 & 97.73 &  35.96 (F1)  & \textit{N/A} & 94.90 & 94.93 \\ 
  
    
    \midrule
     
  &  Human (avg.) & 94.00 & 94.90 & 99.16 & 99.53 & 100  & 98.06 & 95.5 & 95.9 \\  

  &  Human UB & 100 & 100 & 100 & 100 & 100  & 100 & 100 & 100 \\

    \bottomrule
    \end{tabular}
    \caption{Results (\%) of the previous models on the test set of our dataset. \textit{Num} denotes numerical reasoning and \textit{comp} denotes comparison reasoning.
    It is noted that combined reasoning questions require both numerical and comparison reasoning.
    \textit{N/A} denotes not applicable.
   \textit{Human UB} represents the human upper bound.
    }
    \label{tab:result_date}
  \end{center}
\end{table*}

\section{Experiments}

To comprehensively evaluate the top-performing multi-hop models, we conducted various experiments, including both with and without fine-tuning on our dataset.
In addition, to discover the effectiveness of our dataset, we examine the usefulness of our probing tasks and investigate whether our dataset can be used for data augmentation.

\subsection{Models}
As existing models cannot perform all three tasks, we evaluate these models under two groups: one focused on comparison reasoning (e.g., HGN) and the other focused on numerical reasoning (e.g., NumNet+).
HGN~\cite{fang-etal-2020-hierarchical} was designed to deal with HotpotQA, whereas NumNet+~\cite{ran-etal-2019-numnet} was designed to deal with DROP~\cite{dua-etal-2019-drop}.
Both models can perform on the extraction and robustness tasks.
By design, HGN can answer yes/no questions in the comparison reasoning task.
Meanwhile, NumNet+ cannot answer yes/no questions. However, it can deal with numerical questions.
There are some versions of the NumNet model; in our experiment, we use the NumNet+ version.\footnote{\url{https://github.com/llamazing/numnet_plus}}
There are two ways to convert the questions of the extraction task in our dataset to the format of the DROP dataset.
One is to use the span format, and the other is to use the date format.
In our experiment, we use the span format because it produces better results than the date format.

\subsection{Results}
To study the abilities of the models in detail, we evaluate both models on two versions of our dataset: the full-date version (with year-month-date) and the year-only version. We also evaluate the models on two settings: with and without fine-tuning on our dataset.
We use all main and probing questions together for fine-tuning the models.
It is noted that we only use HieraDate when fine-tuning.

\begin{table*}[h]
  \begin{center}
    \begin{tabular}{c c r r r r r r r } 
     \toprule
     
   \multirow{3}{1.0cm}{Model} &   \multirow{3}{2.2cm}{Training Data} &
   \multirow{3}{1.7cm}{\#Questions} & \multicolumn{6}{c}{Evaluation Data}  \\

      \cmidrule{4-9}
   &  &  & \multicolumn{2}{c}{Original}  & Main & Extract & Reason & Robust \\
   \cmidrule{4-9}
   & &  & EM & F1 & F1 & F1 & F1 & F1  \\
       \midrule

 {\multirow{4}{*}{HGN}}  &  Hotpot & 90,447 & 67.56 & 81.13 & 76.25 & 94.64 & 26.03 & 79.74 \\
    
  &  Hotpot \& Ours  & 144,842 & \textbf{67.99} & \textbf{81.44} & \textbf{84.93} & \textbf{97.09} & \textbf{99.95} & \textbf{81.18} \\
 
  
  
     \cmidrule{3-9}
     
  &  2Wiki & 167,454 & 69.42 & 74.21 & 76.69 & 64.62 & \hphantom{0}0.0\hphantom{0} & 77.35 \\
    
  &  2Wiki \& Ours & 221,849 & \textbf{69.66} & \textbf{75.26} & \textbf{85.27} & \textbf{97.03} & \textbf{99.74} & \textbf{82.23} \\
 
 
  

  
 
     \midrule
 {\multirow{2}{*}{NumNet+}}  &  DROP &  77,409 & \textbf{78.58} & \textbf{82.14} & 69.06 & 48.10 & 79.24   & 71.37 \\
    
 &   DROP \& Ours & 120,089 & 78.45 & 82.06 & \textbf{95.39} & \textbf{97.80} & \textbf{94.76} &\textbf{94.54} \\

    
    \bottomrule
    \end{tabular}
    \caption{The results of the HGN and NumNet+ models on HotpotQA, 2Wiki, DROP, and our dataset.
    For the \textit{Original} column, the evaluation data is HotpotQA, 2Wiki, and DROP when the model used HotpotQA, 2Wiki, and DROP for training, respectively.
    %
All reported scores in this table are average scores from two runs.
    }
    \label{tab:result_effective}
  \end{center}
\end{table*}

\paragraph{Date Understanding Evaluation}
Table~\ref{tab:result_date} presents the results of the existing models on the full-date version of our dataset (the year-only version is in Appendix~\ref{datail_year_version}).
When the models are not fine-tuned on our dataset, both HGN and NumNet+ fail in the reasoning task.
This can be because the forms of reasoning questions are new to these models
or the models do not possess the reasoning abilities as humans do.
For the extraction task, HGN performs quite well; meanwhile, NumNet+ performs worse.
In the robustness task, the results are comparable with those of the main multi-hop questions. 
This can be explained by the fact that the patterns of the main multi-hop and robustness questions are similar.

When the models are fine-tuned on our dataset, 
we find that all scores of HGN improve;
especially, HGN reaches the highest score in the comparison reasoning task.
We conjecture that the low scores when HGN is not fine-tuned on our dataset are because the forms of the comparison reasoning questions are new to this model.
Similar to HGN, the scores of the NumNet+ model also improve when it is fine-tuned on our dataset.
However, the score in the numerical reasoning task on the full-date version remains low.
We observed that when we evaluate NumNet+ on the year-only version, the EM scores are 83.1 and 94.4 in the numerical reasoning task for two cases: without and with fine-tuning on our dataset, respectively.
This indicates that NumNet+ could perform subtraction in the form of numbers (as years) but could not in the form of dates.

\paragraph{Dataset Quality Check}
To verify the quality of our dataset, we randomly selected 100 samples from the test set and 
instructed graduate students to conduct the annotation.
Each sample was annotated by two annotators.
We provided the context and a list of questions to the annotators; 
the results are reported in Table~\ref{tab:result_date}.
It can be observed that the human upper bound is 100\% for all tasks.
However, the human average is slightly low. 
On manually investigating the reason for this low human average, 
we found that the annotators made careless mistakes in several examples; however, we confirmed that these examples are answerable and reasonable.
%

\paragraph{Difficulty of Reasoning over Dates}
To discover whether the number of required reasoning skills in each question affects question difficulty, we compared the results of the two main types of questions in our dataset (combined vs. comparison reasoning).
We found that the scores of the comparison reasoning questions were always higher than those of the combined reasoning questions 
(85.7 vs. 72.3 F1 in HGN; 98.8  vs. 81.6 F1 in NumNet+).
The full results are in Appendix~\ref{app_analyses}.
These results indicate that questions requiring both numerical and comparison reasoning are more difficult than questions that require only comparison reasoning.



\paragraph{QA Performance}
To investigate the effectiveness of our probing questions for improving the QA performance, we trained HGN and NumNet+ on six different combinations of the main and probing tasks.
The results show that each task in our dataset helps to improve the performance of the main QA question (all results are in Appendix~\ref{app_sub_quesion_effec}).
Especially when training the models on all tasks, the results improve significantly in both HGN and NumNet+ compared with the models trained on the main questions only (82.7 vs. 77.1 F1 in HGN; 94.9 vs. 84.6 F1 in NumNet+).
This demonstrates that our probing questions not only help to explain the internal reasoning process but also help to improve the score of the main multi-hop questions.

\paragraph{Data Augmentation}
We also check whether our dataset can be used for data augmentation.
We trained HGN and NumNet+ on two settings, on
the original dataset (e.g., HotpotQA) and on the union of the original dataset and our dataset.
We use HGN for HotpotQA and 2Wiki; meanwhile, NumNet+ is used for DROP.
All results are reported in Table~\ref{tab:result_effective}.
There is no significant change on the original datasets (e.g., from 81.1 to 81.4 F1 for HotpotQA); meanwhile, the improvement in our dataset is significant (e.g., from 76.3 to 84.9 F1 on the main QA task).
Notably, all models that are trained on the union of the original dataset and our dataset
are better in our robustness task.
%
This indicates that our dataset can be used as augmentation data for improving the robustness of the models trained on HotpotQA, 2Wiki, and DROP.

\section{Conclusion}

We proposed a new multi-hop RC dataset for comprehensively evaluating the ability of existing models to understand date information.
We evaluated the top-performing models on our dataset.
The results revealed that the models may not possess the ability to subtract two dates even when fine-tuned on our dataset.
We also found that our probing questions could help to improve QA performance, and can be used for data augmentation. 
For future work, we will use the hierarchical manner in our dataset to apply to other types of questions such as numerical reasoning questions in DROP.

\section*{Acknowledgments}
We would like to thank Johannes Mario Meissner Blanco, Napat Thumwanit, and the anonymous reviewers for their comments and suggestions. 
This work was supported by JSPS KAKENHI Grant Numbers 21H03502 and 22K17954.


\bibliography{anthology,custom}
\bibliographystyle{acl_natbib}


\appendix

\section{Limitations}
There are two main limitations in our research.
(1) The proposed dataset, HieraDate, focuses on only the date information.
There is also a lack of diversity of operations in the dataset; it contains only subtraction and comparison operations.
Other operations, such as addition and sorting, are also useful.
We leave the extension for future work.
(2) Another limitation is the results when training the NumNet+ model on the main and robustness tasks (Main \& Robust questions in Table~\ref{tab:result_ablation}).
The results drop significantly, but we have not fully investigated the reasons. 
At this moment, we conjecture that the reason is the contradiction of the two questions in each sample in the training data.
In this setting (Main \& Robust questions), each sample has only two questions, and these two questions are opposite (e.g., ``Who was born first, A or B?'' and ``Who was born later, A or B?'').
This can make the model confused; we will investigate more models on the leaderboard of DROP to find out the reasons.

\section{Dataset Details}
\label{sec_appendix_data}

\subsection{Previous Datasets}
\label{sec_appendix_data_previous}
\paragraph{HotpotQA~\cite{yang-etal-2018-hotpotqa}}
HotpotQA, created through crowdsourcing, includes 
two main types of questions: bridge and comparison.
Unlike previous datasets, a set of sentence-level SFs information is introduced, which facilitates explainable reasoning by the system. 
%
Because of the dataset construction procedure, there is no available information that can be used to generate sub-questions.

\paragraph{2WikiMultiHopQA~\cite{ho-etal-2020-constructing}}
2Wiki was created using Wikipedia articles and Wikidata triples.
Similar to HotpotQA, it includes 
two main types of questions: bridge and comparison.
In 2Wiki, the authors introduced evidence information that can be used to explain the reasoning chain from question to answer.
We used this information for generating sub-questions in our dataset.




%

\subsection{HieraDate Information}
\label{sec_appendix_data_analysis}

\begin{figure}[t]
    \includegraphics[scale=0.6963]{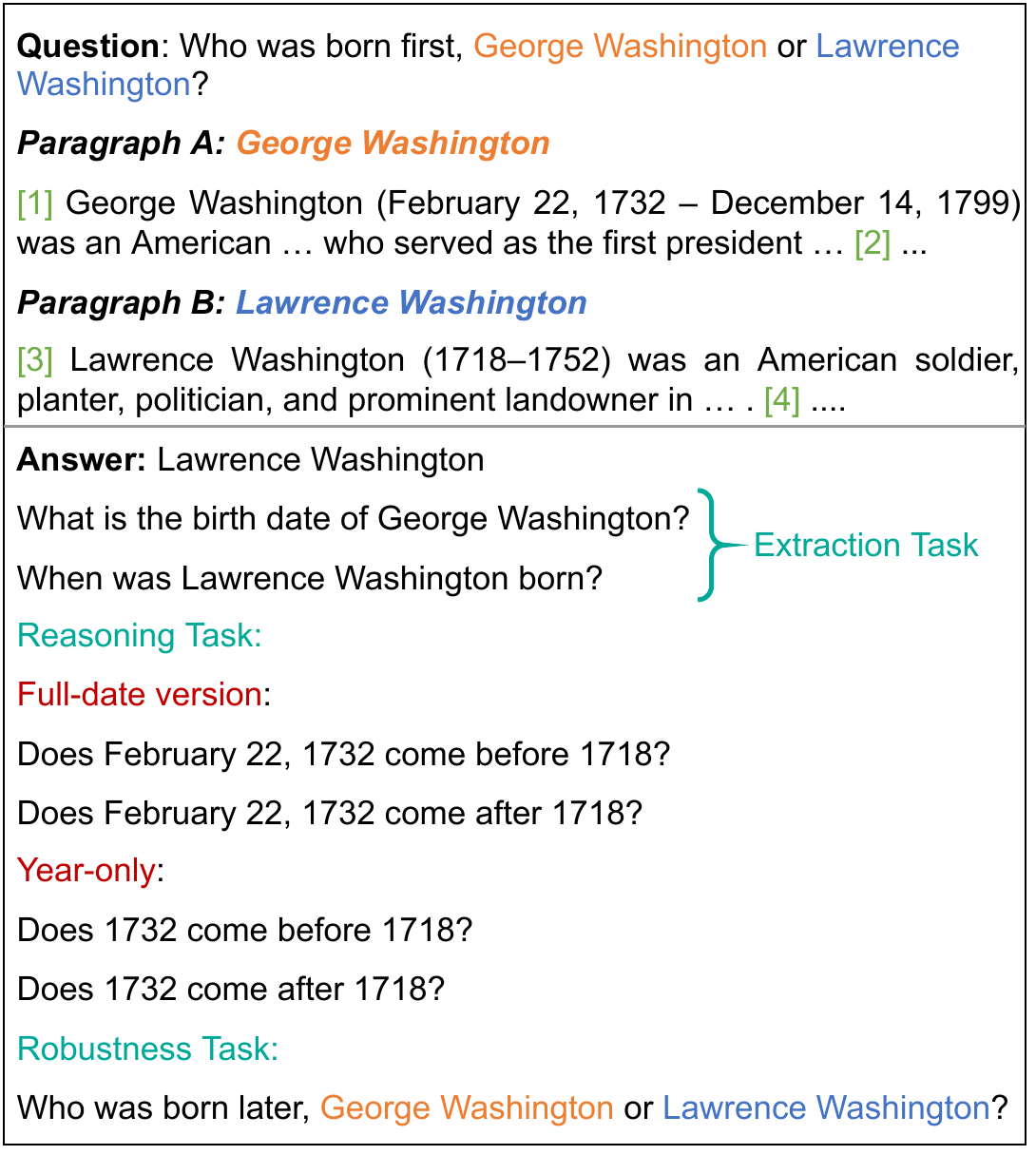}
    \caption{
    Example of a comparison reasoning question in our dataset. 
    }
    \label{fig:example_born}
\end{figure}

\begin{table*}[t]
  \begin{center}
    \begin{tabular}{c l r r r r r r r r} 
     \toprule
     
    \multirow{2}{1.0cm}{Fine-tuning} & \multirow{2}{2cm}{Model} & \multicolumn{2}{c}{Main}  & \multicolumn{2}{c}{Extraction}  &
      \multicolumn{2}{c}{Reasoning} & 
      \multicolumn{2}{c}{Robustness} \\

      \cmidrule{3-10}
     &  & EM & F1 & EM & F1 & EM (num) & EM (comp) & EM & F1 \\
       \midrule

  {\multirow{2}{*}{\blueuncheck}}

   &  HGN & - & - & - & - & \textit{N/A}  & 55.03 & - & - \\

   &  NumNet+ & -  & - & - & - &  83.07   & \textit{N/A} & - & - \\  
    
     \midrule

    {\multirow{2}{*}{\greencheck} }

 &  HGN & 77.23 & 79.24 & 95.84 & 96.93 &  \textit{N/A}  & 99.90 & 76.68 & 78.61 \\ 
    

  &  NumNet+ & 94.90 & 94.93 & 96.29 & 97.69  &  94.36   & \textit{N/A} & 93.99 & 94.01 \\ 
    
        \midrule
  
    {\multirow{2}{*}{\blueuncheck}} 
    
   &  SAE (full) & 69.76 & 77.78 & 82.99 & 84.73 &  \textit{N/A}  & 59.14 & 69.22 & 77.82 \\ 
  &  SAE (year) & - & - & - & - & \textit{N/A}  & 55.75 & - & - \\

    \bottomrule
    \end{tabular}
    \caption{Results (\%) of the previous models on the test set of our dataset. \textit{Num} denotes numerical reasoning and \textit{comp} denotes comparison reasoning.
   %
   %
   ``-'' indicates that the score is similar to the score of the full-date version in the same setting; for HGN and NumNet+, it is similar to the score in Table~\ref{tab:result_date}.
   \textit{N/A} denotes not applicable.
   } 
    \label{tab:result_year_only}
  \end{center}
\end{table*}

\begin{table*}[htp]
  \begin{center}
    \begin{tabular}{l r r r r r r r r} 
     \toprule
     
   \multirow{2}{2cm}{Model-type} & \multicolumn{2}{c}{Main}  & \multicolumn{2}{c}{Extraction}  &
      \multicolumn{2}{c}{Reasoning} & 
      \multicolumn{2}{c}{Robustness} \\

      \cmidrule{2-9}
     & EM & F1 & EM & F1 & EM (num) & EM (comp) & EM & F1 \\
     
       \midrule
    HGN-all & 78.87 & 82.69 & 96.06 & 97.14 &  \textit{N/A}  & 100 & 76.68 & 78.58 \\ 
     HGN-combined & 70.97 & 72.34 & 93.95 & 95.67 &  \textit{N/A}  & 100 & 69.35 & 71.18 \\ 
    
    HGN-comparison & 81.18 & 85.71 & 97.29 & 98.00 &  \textit{NO}  & 100 & 78.82 & 80.74 \\ 
    \cmidrule{2-9}
    \hl{HGN-all} & 75.40 & 76.67 & 95.30 & 96.71 &  \textit{N/A}  & 99.19 & 76.21 & 77.26 \\ 
    
    \hl{HGN-combined} & 66.13 & 67.50 & 93.75 & 95.60  &  \textit{N/A}  & 99.19 & 71.77 & 72.82 \\ 
    
    \hl{HGN-comparison} & 84.68 & 85.85 & 98.39 & 98.92 &  \textit{NO}  & 99.19 & 80.65 & 81.69 \\ 

  \midrule

  NumNet-all & 94.90 & 94.93 & 96.29 & 97.69  &  94.36   & \textit{N/A} & 93.99 & 94.01 \\ 

 NumNet-combined & 81.45 & 81.58 & 94.76 & 96.81  &  94.36   & \textit{N/A} & 79.84 & 79.95 \\ 
  
NumNet-comparison & 98.82 & 98.82 & 97.18 & 98.20  &  \textit{NO}  & \textit{N/A} & 98.12 & 98.12 \\

\cmidrule{2-9}

 \hl{NumNet-all} & 85.08 & 85.43 & 95.97 & 97.69  &  94.00  & \textit{N/A} & 85.48 &  85.50 \\ 

 \hl{NumNet-combined} & 72.58 & 73.27 & 94.76 & 96.84   &  94.00  & \textit{N/A} & 73.39 & 73.42 \\ 
  
\hl{NumNet-comparison} & 97.58  & 97.58 & 98.39 &  99.40 &  \textit{NO}  & \textit{N/A} & 97.58 &  97.58 \\

    \bottomrule
    \end{tabular}
    \caption{Results (\%) of the previous models on the test set of our dataset for different types of questions. 
    \textit{Model-type} denotes the model name and the type of question that the model is evaluated on (e.g., HGN-combined: the results of HGN on combined reasoning questions). 
    \textit{Num} denotes numerical reasoning and \textit{comp} denotes comparison reasoning.
 \textit{N/A} denotes not applicable; meanwhile,
 \textit{NO} indicates that there are no questions for evaluation.
 %
For HGN, we fine-tuned it on the full-date version of our dataset; meanwhile, NumNet+ is fine-tuned on the year-only version of our dataset.
 In the row with highlight color, the model is trained on HieraDate-small where the number of combined reasoning and comparison reasoning questions are equal.
   }
    \label{tab:result_analysis_subtype}
  \end{center}
\end{table*}

\begin{table*}[tp]
  \begin{center}
    \begin{tabular}{l l r r r r r } 
     \toprule
     
 \multirow{2}{1.2cm}{Model} & \multirow{2}{2.5cm}{Training Data} & \multirow{2}{1.7cm}{\#Questions} & \multicolumn{4}{c}{Testing Data}  \\

      \cmidrule{4-7}
  &   &  & Main & Extract & Comp/Num & Robust  \\
       \midrule

  {\multirow{6}{*}{HGN}}  &  Main & 8,745 & 77.11 & \hphantom{0}0.0\hphantom{0} & \hphantom{0}0.0\hphantom{0} & 75.45 \\
    
  &  Main \& Extract & 30,085 & 78.37 & \textbf{97.14} & \hphantom{0}0.0\hphantom{0}  & 78.18 \\
    
  &  Main \& Reason & 24,310 & 79.06 & \hphantom{0}0.0\hphantom{0} & 99.79 & 76.62  \\
    
  &  Main \& Robust & 17,490 & 80.96 & \hphantom{0}0.0\hphantom{0} & \hphantom{0}0.0\hphantom{0} & 78.04  \\
    
  &  Main \& Extract \& Reason & 45,650 & 79.97 & 97.10 & 99.59 & 78.40  \\

  &  All & 54,395 & \textbf{82.69} & \textbf{97.14} & \textbf{100} & \textbf{78.58}   \\
    
    \midrule
    
      {\multirow{8}{*}{NumNet+}}  &  Main & 8,745 & 84.57 & 0.02 & \hphantom{0}0.0\hphantom{0} & 82.87 \\
    
  &  Main \& Extract & 30,085 & 92.03 & 97.75 & \hphantom{0}0.0\hphantom{0}  & 89.28 \\
    
  &  Main \& Reason & 12,595 & 88.92 & 0.19 & 94.36 & 89.83  \\
    
    \cmidrule{3-7}
    
  &  Main \& Robust \#1 & 17,490 & 49.86 & 0.23 & \hphantom{0}0.0\hphantom{0} &  44.84 \\
  
  &  Main \& Robust \#2 & 17,490 & 48.54 & 0.08 & \hphantom{0}0.0\hphantom{0} & 50.42  \\
    
  &  Main \& Robust \#3 & 17,490 & 52.95 & 0.02 & \hphantom{0}0.0\hphantom{0} & 45.24  \\

   \cmidrule{3-7}
    
  &  Main \& Extract \& Reason & 33,935 & 92.01 & \textbf{97.89} &  \textbf{95.16} &  88.91 \\

  &  All & 42,680 & \textbf{94.93} & 97.69 & 94.36 &  \textbf{94.01}  \\

    \bottomrule
    \end{tabular}
    \caption{F1-score of the HGN and NumNet+ models on the test set of our dataset when they are trained on different subsets of our dataset.
\textit{\#Questions} represents the number of questions in the training data.
\textit{Comp/Num} denotes comparison reasoning or numerical reasoning; for the HGN model, it is comparison reasoning; for the NumNet+ model, it is numerical reasoning.
We run three times for the ``Main \& Robust'' setting in the NumNet+ model because the results of this setting are quite different with others.
}
    \label{tab:result_ablation}
  \end{center}
\end{table*}

\paragraph{Question Types}
As mentioned above, there are two main types of questions in our dataset: combined reasoning (Figure~\ref{fig:example_combine}) and comparison reasoning (Figure~\ref{fig:example_born}).
After obtaining all samples from HotpotQA and 2Wiki, 
there are only 11.3\% of combined reasoning questions in the total number of examples.
Therefore, we use Wikidata IDs to retrieve the missing date in a comparison reasoning question to create a combined reasoning question.
For example, if the question asks ``who was born first, Alice or Bob?'', to create a new sample that asks ``who lived longer/shorter'', we need the date of death information.
%
We also have several requirements, such as the date should appear in the paragraph that describes the entity.
After retrieving the missing date, we use the same process as in Section~\ref{data_construction_main} to generate the questions for all three tasks.
It is noted that this converting process is used for the 2Wiki dataset.
In the current version of the dataset, there are 22.1\% of combined reasoning questions.

\paragraph{Date Format}
Wikidata uses a zero value for the dates that miss the month value or day value.
In reality, we have no date with month-0 and day-0; therefore, we use a default value ``1'' for the dates that miss the month value or day value.

\paragraph{Numerical Reasoning Issue}
In reality, in some cases, the paragraph can contain age information, 
e.g., ``\textit{He died in 1981 at the age of 90}''.
In this case, the model does not need to perform numerical reasoning.
We used rules (e.g., filter whether the context contains the word ``age'' or not), then manually checked, and found that there are 13 paragraphs in a total of 248 paragraphs (124 examples) in the test set that the age information is available.


\paragraph{Dataset Versions}
Our dataset has two versions: ``normal setting'' and ``distractor setting''.
The ``normal setting'' includes only two gold paragraphs; meanwhile, the ``distractor setting'' contains ten paragraphs, including two gold paragraphs and eight distractor paragraphs.
In this study, we evaluated the previous models on the ``normal setting''.

\section{Experiments}
\label{sec_appendix_exper}

For NumNet+, we use the parameters as described in the original source code.
%
For HGN, when training it on HotpotQA, 2Wiki, and our dataset, we use only the loss of the answer prediction task.
For other parameters, we use the same parameters as described in the source code\footnote{\url{https://github.com/yuwfan/HGN}} of HGN.

\subsection{Date Understanding Evaluation Details}
\label{datail_year_version}
We also evaluate the previous models on the year-only version of our dataset.
Table~\ref{tab:result_year_only} presents all the results.
When the models are not fine-tuned on our dataset, the score of the HGN model in the comparison reasoning task does not change much when compared with the full-date version (55.0 vs. 53.1 EM); this indicates that there is not much difference between the full-date and year-only versions when using HGN.
For NumNet+, the score of the numerical reasoning task significantly improves when compared with the full-date version (83.1\footnote{In the year-only version, the EM and F1-score are equal.} vs. 22.8 F1); this indicates that NumNet+ can perform numerical reasoning in the form of numbers (as years) but cannot in the form of dates.
%
%

\paragraph{Evaluation on SAE}
Similar to HGN, SAE~\cite{tu-etal-2019-multi} was designed to deal with HotpotQA.
The results are presented in Table~\ref{tab:result_year_only}.
Similar to HGN, the model cannot perform well on the comparison reasoning questions when it is not fine-tuned on our dataset.
As all questions in the comparison reasoning task are yes/no questions,
the random score is 50\%. 
The scores of both HGN and SAE are close to the chance score.

\subsection{Difficulty of Reasoning over Dates}
\label{app_analyses}

Table~\ref{tab:result_analysis_subtype} shows the results of the previous models on the test set of our dataset for different types of questions.
%
As shown in the table, the scores of comparison reasoning questions were always higher than those of combined reasoning questions.
In the current version of the dataset, there are only 22.1\% combined reasoning questions.
To avoid the data-size bias, we created a HieraDate-small version by randomly choosing the comparison reasoning questions such that the number of combined reasoning questions is equal to the number of comparison reasoning questions.
We then conducted experiments on HieraDate-small.
We found similar results as on HieraDate.
These results indicate that combined reasoning questions are more difficult than comparison reasoning questions.

\subsection{QA Performance}
\label{app_sub_quesion_effec}
Table~\ref{tab:result_ablation} presents the results of the HGN and NumNet+ models on the test set of our dataset when they are trained on different subsets of our dataset.

\subsection{Error Cases}
Table~\ref{tab_error_case} presents some error cases of the previous models on the test set of our dataset.

\begin{table*}[t]
\centering
\begin{tabular}{p{5cm}p{3.25cm}p{6.5cm}}
\toprule
Context & Main question  & Sub-questions \\ 

\midrule

\multirow{9}{5cm}{\textbf{Paragraph A:} {\color{orange} Lotte Backes} (May 2, 1901 - May 12, 1990) was a German pianist, \ldots \\ 
\textbf{Paragraph B:} {\color{blue} Willem van Haecht} (1593 – 12 July 1637) was a Flemish painter best known for his pictures \ldots} 
& \multirow{9}{3.25cm}{\textbf{Q:} Who died first, {\color{orange} Lotte Backes} or {\color{blue} Willem van Haecht}? \\
 \textbf{Predicted answer:} Willem van Haecht \greencheck}
&  \multirow{9}{6.5cm}{\textbf{Q1:} Does May 12, 1990 come before July 12, 1637? \\ \textbf{Predicted 1:} yes \reduncheck \\   \textbf{Q2:} Does May 12, 1990 come after July 12, 1637? \\ \textbf{Predicted 2:} yes \greencheck \\
\textbf{Year-only version:} \\
\textbf{Q3:} Does 1990 come before 1637? \\ \textbf{Predicted 3:} yes \reduncheck \\   \textbf{Q4:} Does 1990 come after 1637? \\ \textbf{Predicted 4:} yes \greencheck
} 

\\  \\ \\ \\ \\ \\  \\ \\ \\ \\ \midrule


\multirow{12}{5cm}{\textbf{Paragraph A:} {\color{orange} Andrzej Markowski} (22 August 1924 – 30 October 1986) was a Polish composer and conductor. \ldots \\ 
\textbf{Paragraph B:} {\color{blue} François Missoffe} (13 October 1919 in Toulon, France – 28 August 2003 in Rouen) was a French politician and diplomat. \ldots} 
& \multirow{12}{3.25cm}{\textbf{Q:} Who lived longer, {\color{orange} Andrzej Markowski} or {\color{blue} François Missoffe}? \\
 \textbf{Predicted answer:} François Missoffe \greencheck}
&  \multirow{12}{6.5cm}{\textbf{Q1:} How old was Andrzej Markowski when they died? \\ \textbf{Predicted 1:} 62 (number format) \greencheck \\   \textbf{Q2:} How old was François Missoffe when they died? \\ \textbf{Predicted 2:} 84 (number format) \greencheck \\ \textbf{Q3:} Is a 62-year-2-month-8-day-old person older than a 83-year-10-month-15-day-old person? \\ \textbf{Predicted 3:} yes \reduncheck \\
\textbf{Q4:} Is a 62-year-old person older than a 83-year-old person? \\ \textbf{Predicted 4:} yes \reduncheck (year-only version) } 
\\  \\ \\ \\ \\ \\ \\ \\ \\ \\ \\ \\ \\ \midrule


\multirow{12}{5cm}{\textbf{Paragraph A:} {\color{orange} Oliver A. Unger} (August 28, 1914 – March 27, 1981) was an award- winning American film producer, distributor, \ldots \\
\textbf{Paragraph B:} {\color{blue} Ross Story} (16 January 1920 – 9 May 1991), always known as Ross or C. R. Story, was a farmer and politician \ldots} 
& \multirow{12}{3.25cm}{\textbf{Q:} Who died later, {\color{orange} Oliver A. Unger} or {\color{blue} Ross Story}? \\
 \textbf{Predicted answer:} Ross Story \greencheck}
&  \multirow{12}{6.5cm}{\textbf{Q1:} What is the death date of Oliver A. Unger? \\ \textbf{Predicted 1:} 9 May 1991 \reduncheck \\   \textbf{Q2:} What's the death date of Ross Story? \\ \textbf{Predicted 2:} 9 May 1991 \greencheck \\ \textbf{Q3:} Does March 27, 1981 come before May 09, 1991? \\ \textbf{Predicted 3:} yes \greencheck \\ \textbf{Q4:} Does March 27, 1981 come after May 09, 1991? \\ \textbf{Predicted 4:} no \greencheck}
\\  \\ \\ \\ \\ \\ \\ \\ \\ \\ \\ \\

\bottomrule
\end{tabular}
\caption{\label{tab_error_case} Error cases of the previous models on our dataset (without fine-tuning).
It is noted that there are no existing models that can perform all three tasks. The results in example \#2 are from the two models, HGN and NumNet+.
In the first example, we can see that the models do not have the ability to compare two dates or two years.
In example \#2, we can observe that the models do not have the ability to subtract two dates, but the models can calculate the age by simply subtracting two years of the two dates.
In example \#3, we observe that the models can answer the main multi-hop question correctly, although they do not know what the date of death of a person is.
}

\end{table*}


\end{document}